\documentclass[runningheads]{llncs}

\usepackage{accv}

\usepackage{accvabbrv}

\usepackage{graphicx}
\usepackage{booktabs}

\usepackage[accsupp]{axessibility}  

\usepackage{amsmath}
\usepackage{multirow} 
\usepackage{graphicx} 
\usepackage{xcolor} 
\usepackage{rotating} 
\usepackage{graphicx,xcolor}
\usepackage{pxrubrica}
\usepackage{url}
\usepackage{bm}
\usepackage{amssymb}
\usepackage{cite}
\usepackage{subcaption}
\usepackage{graphicx}
\usepackage{arydshln}
\usepackage{amsmath}
\usepackage{amssymb}
\usepackage{booktabs}
\usepackage{tabularx}
\usepackage{bm}
\usepackage{cite}
\usepackage{float}
\usepackage{wrapfig}
\usepackage{calc}
\PassOptionsToPackage{table}{xcolor}
\usepackage{multirow}
\usepackage{tikz}
\usetikzlibrary{positioning}
\usepackage {colortbl,array,xcolor}
\definecolor{testcolor}{rgb}{1, 1, 0.6}
\definecolor{myred}{HTML}{B85450}
\definecolor{myblue}{HTML}{6C8EBF}
\definecolor{myorig}{HTML}{82B366}
\definecolor{myfoil}{HTML}{C787C9}
\definecolor{gray}{HTML}{818589}
\definecolor{mypurple}{RGB}{211,91,255}
\definecolor{mygreen}{RGB}{18,191,39}
\definecolor{polaris}{RGB}{121,150,196}
\definecolor{composite}{RGB}{230,163,126}
\definecolor{flickr8k-cf}{RGB}{129,190,142}
\definecolor{flickr8k-ex}{RGB}{211,123,126}
\definecolor{TitleColor}{gray}{0.95}
\definecolor{TitleColor}{rgb}{208, 53, 136}
\definecolor{LightCyan}{rgb}{0.88,0.95,1}

\usepackage{hyperref}

\usepackage{orcidlink}

\begin{document}

\title{\textsc{Deneb}: A Hallucination-Robust Automatic Evaluation Metric for Image Captioning} 
\author{Kazuki Matsuda \and Yuiga Wada \and Komei Sugiura}
\institute{Keio University, Japan\\\email{\{k2matsuda0, yuiga, komei.sugiura\}@keio.jp}}
\authorrunning{K.~Matsuda et al.}
\titlerunning{\textsc{Deneb}}

\maketitle
\begin{abstract}
In this work, we address the challenge of developing automatic evaluation metrics for image captioning, with a particular focus on robustness against hallucinations.
Existing metrics are often inadequate for handling hallucinations, primarily due to their limited ability to compare candidate captions with multifaceted reference captions. 
To address this shortcoming, we propose \textsc{Deneb}, a novel supervised automatic evaluation metric specifically robust against hallucinations. 
\textsc{Deneb} incorporates the Sim-Vec Transformer, a mechanism that processes multiple references simultaneously, thereby efficiently capturing the similarity between an image, a candidate caption, and reference captions.
To train \textsc{Deneb}, we construct the diverse and balanced Nebula dataset comprising 32,978 images, paired with human judgments provided by 805 annotators.
We demonstrated that \textsc{Deneb} achieves state-of-the-art performance among existing LLM-free metrics on the FOIL, Composite, Flickr8K-Expert, Flickr8K-CF, Nebula, and PASCAL-50S datasets, validating its effectiveness and robustness against hallucinations.
Project page at \url{https://deneb-project-page-nc03k.kinsta.page/}.
\end{abstract}
\section{Introduction}
\label{sec:intro}

Image captioning has been extensively researched and applied in various social applications, such as the assistance of visually impaired individuals, the analysis of medical images, and the generation of explanations in robotics \cite{gurari2020captioning, ahsan2021IC, ghandi2022deep, pavlopoulos-etal-2019-survey, AYESHA2021107856, kambara2022future, ogura2020attnbr}.
 In scenarios where `AI safety' is paramount, generating appropriate and reliable captions is crucial to avoiding the misrepresentation of the content of images. 
In particular, erroneous captions that include words not depicted in the image, commonly referred to as `hallucinations', are a prevalent issue in image captioning \cite{Rohrbach2018Objecthal, FOIL}.  
However, a significant issue exists in current image captioning research: models failing to address hallucinations are often wrongfully overrated, as existing evaluation metrics predominantly focus only on correlation with human judgment, and overlook critical flaws.
This misalignment is especially problematic in the context of social applications where reliability is essential.
Despite their importance, most existing metrics inadequately address the issue of hallucinations.
In fact, some studies have demonstrated that while most \textit{data-driven} metrics \cite{clipscore, bertscore, moverscore, vilbertscore, mid} correlate well with human judgments, they are less effective in addressing hallucinations \cite{cider, clipscore, pac-s, polos}.

Hallucination is a prevalent issue in text generation models, including Large Language Models (LLMs)\cite{manakul2023selfcheckgpt, mishra2024finegrained, krishna2017visualgenome, sun2023aligning, li2023evalhal, furkan2022hal, FOIL}. 
Despite their rapid advancement and adoption in various societal applications, LLMs often generate hallucinated text.
This suggests that most LLMs are internally unable to evaluate hallucinations and that assessing hallucinations is a particularly challenging task.
The FOIL benchmark \cite{FOIL} has been introduced to evaluate the robustness of metrics against hallucinations in image captioning. 
Several studies \cite{clipscore, pac-s, mid, polos} have revealed that most metrics still fall short of human performance on this benchmark, despite their strong correlation with human judgments.

\begin{figure}[t]
    \centering
    \includegraphics[width=1.0\linewidth]{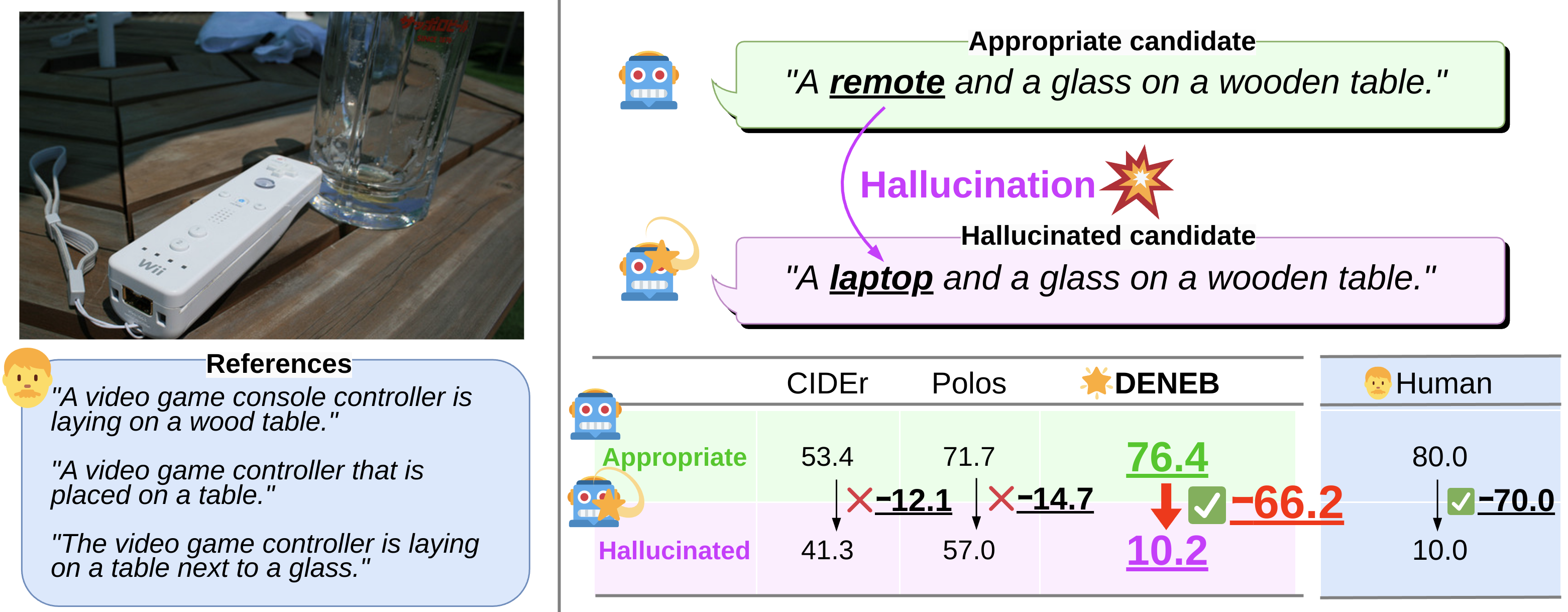}
    \caption{\textbf{Overview of \textsc{Deneb}}. Our metric is designed to effectively evaluate hallucinated captions, which is crucial in scenarios where `AI safety' is paramount. Unlike existing metrics such as CIDEr \cite{cider} and Polos \cite{polos}, which often fail to distinguish between correct and hallucinated captions, \textsc{Deneb} demonstrates improved robustness by assigning lower scores to hallucinated captions than correct captions.}
    \label{fig:eye-catch}
\end{figure}

Automatic evaluation metrics for image captioning can be broadly classified into four categories: \textit{classic} metrics, \textit{reference-free} metrics, \textit{pseudo-multifaceted} metrics, and \textit{multifaceted} metrics.
\textit{Classic} metrics\cite{bleu, meteor, cider, rouge, spice, jaspice} are traditional approaches primarily based on $n$-grams and/or scene graphs. 
Despite their widespread use, they often provide low correlation with human judgments, leading to the development of \textit{data-driven} metrics such as \textit{reference-free} and \textit{pseudo-multifaceted} metrics.
\textit{Reference-free} metrics \cite{clipscore, pac-s, umic}, which use images and do not rely on reference captions, correlate better with human judgments than \textit{classic} metrics.  
However, they face challenges in robustness against hallucinations. 
This issue mainly stems from their heavy dependence on the alignment between image and language features, which has shown to be insufficiently effective in accurately grounding specific local areas within images.  
In fact, some studies have indicated that even the classic metric CIDEr \cite{cider} outperforms CLIP-S in terms of robustness against hallucinations \cite{clipscore}.

In contrast to \textit{classic} and \textit{reference-free} metrics, 
\textit{pseudo-multifaceted} metrics \cite{clipscore, pac-s, polos, mid} utilize both image and text encoders to process an image and multiple reference sentences. 
These metrics outperform \textit{reference-free} metrics in terms of correlation with human evaluations but have significant issues in effectively leveraging multiple reference sentences. 
In general, they compute evaluation scores for each reference sentence independently, resulting in the virtual processing of only a single reference sentence and leading to suboptimal performance.

Handling multiple references becomes particularly important in the automatic evaluation of image captioning.
Consider a scenario where an image depicts both a person and a dog, with two corresponding references. The first reference describes only the person, while the second focuses solely on the dog.
Previous works, by utilizing an aggregation mechanism, virtually processes only a single reference. 
Therefore, they fail to evaluate a candidate that properly describes both subjects, such as ``a person walking a dog.’’
This underscores the necessity of handling multiple references.

Therefore, we propose \textsc{Deneb}\footnote{In the field of general text generation metrics, names of celestial bodies are often used as names for metrics, such as METEOR\cite{meteor} and COMET\cite{comet}.
Following this convention, we have named our metric and dataset after a celestial body.}, a supervised automatic evaluation metric specifically robust against hallucinations. 
Fig. \ref{fig:eye-catch} shows an overview of \textsc{Deneb}.
Unlike \textit{pseudo-multifaceted} metrics, \textsc{Deneb} is a \textit{multifaceted} metric and thus can effectively compare multifaceted descriptions of an image against a candidate caption.
This capability is achieved by the Sim-Vec Transformer, a novel architecture designed for the efficient processing of multiple reference captions.

Furthermore, for training our proposed metric, we construct the Nebula dataset by extending the Polaris dataset, increasing the visual diversity by a factor of three.
Our dataset offers more effective data for training supervised metrics, particularly those capable of handling a wide range of cases, including hallucinated captions.

The main contributions of this study are summarized as follows:
\begin{enumerate}
    \setlength{\parskip}{0.5mm}
    \setlength{\itemsep}{0.2mm}
    \item We introduce the Sim-Vec Transformer, which handles the similarity between an image, a candidate caption, and reference captions.
    \item We propose Sim-Vec Extraction (SVE), which utilizes a Hadamard product and element-wise differences to extract features beneficial for evaluation.
    \item We construct the diverse and balanced Nebula dataset comprising 32,978 images, paired with human judgments provided by 805 annotators.
    \item We achieve state-of-the-art performance among LLM-free metrics on FOIL, Composite, Flickr8K-Expert, Flickr8K-CF, PASCAL-50S, and Nebula.
\end{enumerate}

\section{Related Work}
\label{sec:related}
Image captioning has been widely applied in various areas of society, including the assistance of visually impaired individuals\cite{gurari2020captioning, ahsan2021IC, dognin2022image, ghandi2022deep}, medical image analysis\cite{pavlopoulos-etal-2019-survey, AYESHA2021107856}, and the generation of explanations in robotics\cite{kambara2022future, ogura2020attnbr}.
Several comprehensive surveys \cite{hossain2019surveya, stefanini2023surveyb, ghandi2022deep} have provided an exhaustive overview of representative models, standard datasets, and metrics, including those specifically designed for evaluating hallucinations, such as CHAIR \cite{Rohrbach2018Objecthal}. 
Notably, \cite{ghandi2022deep} offers a detailed summary of how different approaches, including convolutional methods, attention mechanisms, and generative adversarial networks, contribute to reducing hallucinations in image captioning.

\paragraph{\textbf{\textrm{Hallucinations.}}} 
Hallucinations pose significant challenges not only in LLMs but also in image captioning models \cite{manakul2023selfcheckgpt, mishra2024finegrained, krishna2017visualgenome, sun2023aligning, li2023evalhal, furkan2022hal, FOIL}.
In image captioning, hallucinations are instances in which models generate captions with words not corresponding to any of the elements in the input image. 
As noted in \cite{ghandi2022deep}, this issue is particularly critical in social applications where the correctness of captions is prioritized over content coverage.
Such hallucinations substantially affect the reliability of these models.

A notable metric developed to evaluate this issue is CHAIR \cite{Rohrbach2018Objecthal}, which assesses hallucinations by calculating the proportion of generated words that accurately reflect objects in the input image.
 However, CHAIR's approach is rule-based and operates within a closed vocabulary, limiting its applicability and generalizability. 
The development of CHAIR underscores the need for more comprehensive benchmarks to evaluate the robustness against hallucinations.

In response to the challenge of hallucinations, Shekhar et al. \cite{FOIL} introduced the FOIL benchmark to evaluate the robustness of metrics against hallucinations. The FOIL dataset, derived from the COCO dataset \cite{coco}, consists of approximately 200,000 image-caption pairs, featuring a mix of both correct and hallucinated captions. Hallucinated captions are generated by altering a single word in a correct caption to create a similar but inaccurate version — for instance, replacing ``motorcycle'' with ``bicycle''.
Moreover, Hessel et al. \cite{clipscore} presented a widely-accepted methodology for using the FOIL dataset to evaluate metrics. As detailed in \cite{clipscore}, this methodology entails the following steps: For each of the 32K test images in the dataset, a pair comprising a hallucinated caption and its correct counterpart is sampled. The robustness of each evaluation metric against hallucinations is then evaluated based on its capability to consistently assign higher scores to the correct captions than to their hallucinated versions.

\paragraph{\textrm{\textbf{Image Captioning Metrics.}}}  
Standard metrics for image captioning include BLEU \cite{bleu}, METEOR \cite{meteor}, ROUGE \cite{rouge}, CIDEr \cite{cider, cider-r}, and SPICE \cite{spice, jaspice}.
These \textit{classic} metrics, primarily based on $n$-grams and/or scene graphs, have been extensively used but often exhibit low correlation with human judgments. This discrepancy has led recent studies to shift their focus toward \textit{data-driven} metrics, such as BERTScore \cite{bertscore}, CLIP-S \cite{clipscore}, MID \cite{mid}, and Polos \cite{polos}. 
BERTScore and MoverScore, for instance, leverage a pre-trained BERT encoder to compare word token embeddings in both candidate captions and references. However, their lack of image incorporation can limit their effectiveness in evaluating image captioning models.

Several metrics adopt strategies that leverage both visual and language embeddings from pre-trained vision-and-language models, such as CLIP \cite{clip}, ViLBERT \cite{vilbert}, and UNITER \cite{uniter}.
A notable example is CLIP-S \cite{clipscore}, which employs an unsupervised approach to evaluate captions by measuring their similarity to embeddings generated by CLIP encoders. 
The distinct feature of CLIP-S is its ability to evaluate captions in contexts both with and without reference images.
In \cite{pac-s}, the authors introduced PAC-S, a variant of CLIP-S. In this variant, the authors generated text-image pairs using image generators and the original CLIP model was fine-tuned with these generated pairs.

Supervised metrics, which are trained based on human judgments, have been studied in text generation tasks, including machine translation, text summarization, and text simplification \cite{comet, ruse, bleurt, maddela-etal-2023-lens, xu-etal-2023-sescore2}.
In the field of image captioning, however, the development of such metrics is still in its infancy. 
UMIC is an example of a supervised metric in this field and employs the fine-tuning of UNITER through a ranking-based approach. 
While UMIC \cite{umic} has shown superior performance to other metrics,  \cite{polos} pointed out that ranking models such as UMIC have shortcomings, such as varying focal points in captions and subjective variations in expression.
A notable instance of a different approach from the ranking model is Polos,  a supervised metric inspired by COMET \cite{comet}, BLEURT \cite{bleurt}, and RUSE \cite{ruse}. 
The authors of \cite{polos} introduced the $\mathrm{M^2LHF}$ framework, which was used to develop a metric utilizing human judgments.
Based on the $\mathrm{M^2LHF}$, Polos achieves state-of-the-art performance in various image captioning benchmarks. 
However, as outlined in Section \ref{sec:intro}, Polos falls into the category of \textit{pseudo-multifaceted} metrics, facing challenges in effectively managing multiple references.

\paragraph{\textbf{\textrm{Datasets and Benchmarks.}}}
Standard datasets for evaluating image captioning metrics include FOIL \cite{FOIL}, Flickr8K-Expert \cite{flickr}, Flickr8K-CF, Composite \cite{composite}, Polaris \cite{polos}, and PASCAL-50S \cite{spice}.
In response to concerns about the lack of diversity in the Flickr8K-Expert, Flickr8K-CF, and Composite datasets, Wada et al. \cite{polos} introduced the Polaris dataset, which is currently the largest in the field. 
However, as we argue later, Polaris faces the challenge of a limited variety of images compared with the total number of samples, leading to concerns about imbalance in the dataset.

\section{Methods}
\label{sec:method}

\begin{figure}[t]
    \centering
    \includegraphics[width=0.95\linewidth]{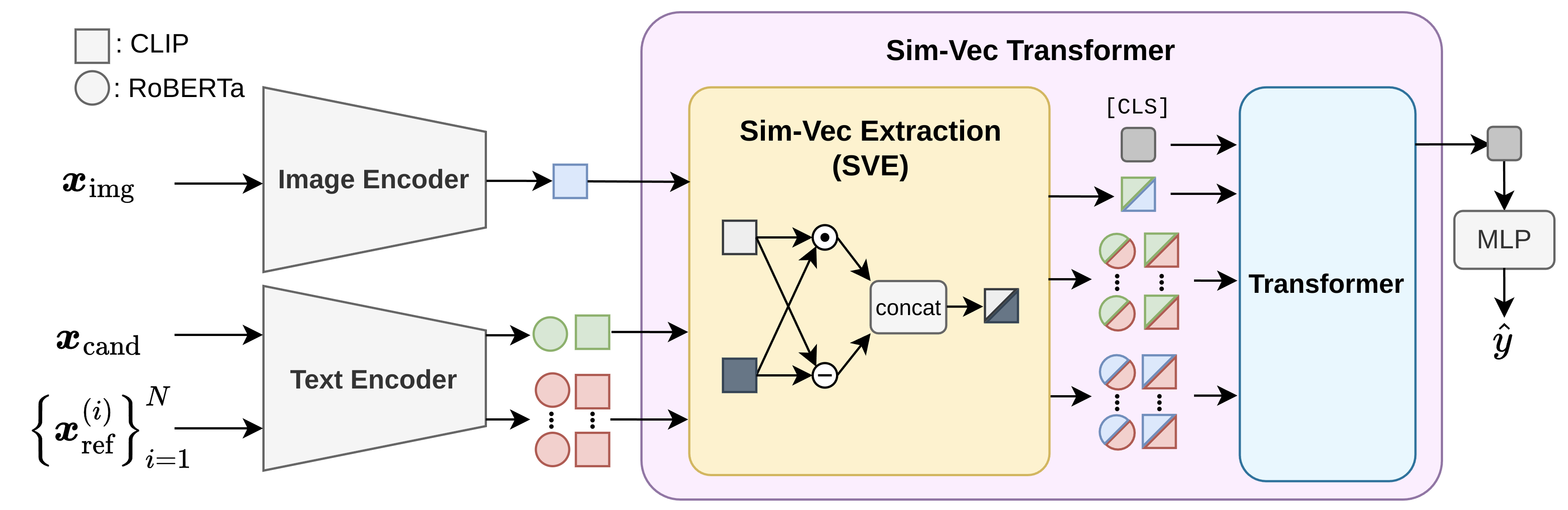}
    \caption{\textbf{The architecture of \textsc{Deneb}}. CLIP and RoBERTa are employed to extract embeddings from an image, a candidate, and references. These embeddings are then processed concurrently by the Sim-Vec Transformer, comprising two modules: the Sim-Vec Extraction and transformer. Sim-Vec Extraction (SVE) utilizes a Hadamard product and element-wise differences to extract features, capturing the similarity among $\bm{x}_\mathrm{img}$, $\bm{x}_\mathrm{cand}$, and $\{\bm{x}_\mathrm{ref}^{(i)}\}_{i=1}^N$.}
    \label{fig:network}

\end{figure}

In this study, we propose a novel automatic evaluation metric \textsc{Deneb}, which is specifically robust to hallucinations.
Fig. \ref{fig:network} provides an overview of the proposed metric \textsc{Deneb}.
Unlike \textit{pseudo-multifaceted} metrics, \textsc{Deneb} is a \textit{multifaceted} metric and can therefore effectively compare multifaceted descriptions of an image with a candidate caption.
This capability is achieved by the Sim-Vec Transformer, a novel architecture designed for the efficient processing of multiple reference captions.

The proposed metric distinguishes itself from existing metrics in three key aspects as follows:
First, in contrast to \textit{classic} metrics\cite{cider, bleu, rouge, meteor}, \textsc{Deneb} is a \textit{data-driven} automatic evaluation metric that employs a supervised approach. 
Second, unlike \textit{reference-free} metrics \cite{clipscore, pac-s, umic}, \textsc{Deneb} utilizes reference captions, enhancing its robustness against hallucinations.
Finally, whereas \textit{pseudo-multifaceted} metrics typically employ aggregate functions \cite{clipscore, pac-s, polos}, \textsc{Deneb} more effectively addresses the challenge of hallucination by handling multiple references simultaneously through the Sim-Vec Transformer.

The Sim-Vec Transformer handles the similarity between an image, a candidate caption, and reference captions.
This module can be broadly applied to automatic evaluation metrics that deal with image features, including CLIP-S \cite{clipscore} and PAC-S \cite{pac-s}.

\subsection{Feature Extraction}

Given an image $\bm{x}_\mathrm{img}$, a candidate $\bm{x}_\mathrm{cand}$, and $N$ references $\{\bm{x}_\mathrm{ref}^{(i)}\}_{i=1}^N$ , the automatic evaluation metrics for image captioning models output a score $\hat{y}$ that captures the appropriateness of $\bm{x}_\mathrm{cand}$ in the context of $\bm{x}_\mathrm{img}$ and $\{\bm{x}_\mathrm{ref}^{(i)}\}_{i=1}^N$.
First, the input $x$ to our metric is defined as follows:
\begin{align}
\bm{x} = \left\{\bm{x}_\mathrm{img}, \left\{\bm{x}_\mathrm{ref}^{(i)}\right\}_{i=1}^N , \bm{x}_\mathrm{cand}\right\},
\end{align}
where $\bm{x}_\mathrm{img} \in \mathbb{R}^{3 \times H \times W}$, $\{\bm{x}_\mathrm{ref}^{(i)}\}_{i=1}^N \in \{0, 1\}^{N \times V \times L}$, and $\bm{x}_\mathrm{cand} \in \{0, 1\}^{V \times L}$ represent an image, $N$ references, and a candidate, respectively.
Furthermore, $H$, $W$, $N$, $V$, $L$ denote the height and width of the image, the number of references, the vocabulary size, and the number of tokens, respectively.

In this study, we extract features from $\bm{x}$ using CLIP \cite{clip} and RoBERTa\cite{RoBERTa}.
We utilize the CLIP encoder (CLIP ViT-B/32) pre-trained with \cite{pac-s} and RoBERTa-base pre-trained with SimCSE\cite{simcse}.
Specifically, we use CLIP's text encoder to obtain text embeddings, $\{\bm{r}_\mathrm{clip}^{(i)}\}_{i=1}^N \in \mathbb{R}^{N \times d_\mathrm{clip}}$ and $\bm{c}_\mathrm{clip} \in \mathbb{R}^{d_\mathrm{clip}}$, from $\{\bm{x}_\mathrm{ref}^{(i)}\}_{i=1}^N$ and $\bm{x}_\mathrm{cand}$, respectively, where $d_\mathrm{clip}$ denotes the output dimension of CLIP. 
Concurrently, using the CLIP image encoder, we obtain image features $\bm{v} \in \mathbb{R}^{d_\mathrm{clip}}$ from $\bm{x}_\mathrm{img}$. 
Additionally, we employ RoBERTa to obtain text embeddings $\{\bm{r}_\mathrm{rb}^{(i)}\}_{i=1}^N \in \mathbb{R}^{N \times d_\mathrm{rb}}$ and $\bm{c}_\mathrm{rb} \in \mathbb{R}^{d_\mathrm{rb}}$ from $\{\bm{x}_\mathrm{ref}^{(i)}\}_{i=1}^N$ and $\bm{x}_\mathrm{cand}$, respectively, where $d_\mathrm{rb}$ denotes the  output dimension of RoBERTa and $\{\bm{r}_\mathrm{rb}^{(i)}\}_{i=1}^N$ and $\bm{c}_\mathrm{rb}$ are obtained from the \verb|[CLS]| token of the inputs.
In this study, both the CLIP and RoBERTa encoders are frozen.

\subsection{Sim-Vec Transformer}

\paragraph{\textbf{Sim-Vec Extraction.}}
In the context of metrics for image captioning, accurately capturing the similarity between the image, the candidate, and the multiple references is crucial. 
Several studies have incorporated mechanisms to extract these similarities, thereby achieving high performance\cite{polos, comet, ruse, maddela-etal-2023-lens}.
According to the observations in \cite{polos, comet, ruse, maddela-etal-2023-lens}, the exclusion of the candidate and the reference leads to a decline in performance.
This suggests that these metrics do not fully leverage the extracted vector-form similarities, partially because the latter part of them, specifically responsible for processing vector-form similarities, is just a poorly performing MLP. 
Considering these mechanisms effectively extract similarities  \cite{polos, comet, ruse, maddela-etal-2023-lens}, the inclusion of raw features of candidates and references could even hinder effective learning from vector-form similarities.
Therefore, in this study, we introduce the Sim-Vec Extraction (SVE) module by extending them to enhance the focus on effectively leveraging these extracted similarities.
Our transformer-based metric has greater potential in capturing vector-form similarities more effectively than the classic MLP approach. 

The SVE module employs a Hadamard product and element-wise differences to extract features that capture the similarity from the following vectors:
\begin{align}
\label{eq:2}
    \bm{s}_\mathrm{in} = \left\{ \bm{c}_\mathrm{clip},  \left\{ \bm{r}_\mathrm{clip}^{(i)}\right\}_{i=1}^{N}, \bm{c}_\mathrm{rb},  \left\{\bm{r}_\mathrm{rb}^{(i)}\right\}_{i=1}^{N} , \bm{v}\right\}.
\end{align}
We compute $\bm{h}_\mathrm{clip}$, $\bm{h}_\mathrm{rb}$, $\bm{d}_\mathrm{clip}$, $\bm{d}_\mathrm{rb}$ as follows:
\begin{align}
\label{eq:3}
\bm{h}_\mathrm{clip} &= \left\{
    \bm{c}_\mathrm{clip}\odot\bm{v},
    \left\{
    \bm{c}_\mathrm{clip}\odot\bm{r}_\mathrm{clip}^{(i)}
    \right\}_{i=1}^{N}
\right\}, \\
\label{eq:4}
\bm{d}_\mathrm{clip} &= \left\{
    |\bm{c}_\mathrm{clip}-\bm{v}|,
    \left\{
    |\bm{c}_\mathrm{clip}-\bm{r}_\mathrm{clip}^{(i)}|
    \right\}_{i=1}^{N}
\right\}, \\
\label{eq:5}
\bm{h}_\mathrm{rb} &= \left\{
    \bm{c}_\mathrm{rb}\odot\bm{r}_\mathrm{rb}^{(i)}
\right\}_{i=1}^{N}, \\
\label{eq:6}
\bm{d}_\mathrm{rb} &= \left\{
    |\bm{c}_\mathrm{rb}-\bm{r}_\mathrm{rb}^{(i)}|
\right\}_{i=1}^{N}.
\end{align}
It is important to note that, unlike RUSE \cite{ruse} and COMET \cite{comet}, we do not combine the candidate features $\bm{c}$ and the 
 reference features $\bm{r}$ here to allow our model to focus more on vector-form similarity. 
Subsequently, we obtain $\bm{g}_\mathrm{inter}$ by combining $\bm{h}_\mathrm{clip}$, $\bm{d}_\mathrm{clip}$, $\bm{h}_\mathrm{rb}$, and $\bm{d}_\mathrm{rb}$.
\begin{align}
        &\bm{g}_\mathrm{inter} = \left\{
        \bm{h}_\mathrm{clip},
        \bm{d}_\mathrm{clip},
        \bm{h}_\mathrm{rb},
        \bm{d}_\mathrm{rb}
    \right\}.
\end{align}
\paragraph{\textbf{Non-aggregate mechanism.}}
Previous studies \cite{polos, pac-s, clipscore} exhibit a fundamental limitation in the handling of multiple references.
Specifically, they calculate evaluation scores independently for each of the $N$ references, so that the multiple references are not fully utilized.
Generally, CLIP-S\cite{clipscore}, PAC-S\cite{pac-s}, and Polos\cite{polos} can be formulated as follows:
\begin{align}
    \hat{y} = \underset{i}{\mathrm{Aggregate}} \ f(\bm{x}^{(i)}),
\end{align}
where $\bm{x}^{(i)}$ denotes the $i$th reference included in $\bm{x}^{(i)}=\{\bm{x}_\mathrm{img}, \bm{x}_\mathrm{ref}^{(i)}, \bm{x}_\mathrm{cand}\}$ and $\mathrm{Aggregate}\:(\cdot)$ represents any mapping $f: \mathbb{R}^N \rightarrow \mathbb{R}$, such as the Max and Mean functions.
In these methods, the core of the metric $f(\cdot)$ takes only a single reference, $\bm{x}_\mathrm{ref}^{(i)}$, as input, implying the ineffective processing of multiple references.
Moreover, the aggregate function cannot be optimized since it does not have trainable parameters.
Therefore, it can be concluded that these methods do not fully utilize multiple references in the automatic evaluation of image captioning.

To overcome this limitation, \textsc{Deneb} employs an $N$-layer transformer to effectively handle multiple references.
Initially, we obtain $\bm{g}$ by concatenating the \verb|[CLS]| token $\bm{g}_{\texttt{[CLS]}}$ to $\bm{g}_\mathrm{inter}$ obtained from the SVE as follows:
\begin{align}
    \bm{g}=\left\{\bm{g}_{\texttt{[CLS]}},\bm{g}_\mathrm{inter}\right\}. 
\end{align}
Here, $\bm{g}_\mathrm{inter}$ is defined as the feature extracted from the SVE. 
Subsequently, we calculate $\bm{h}_\mathrm{N}$ using an $N$-layer transformer encoder:
\begin{align}
    \bm{h}_\mathrm{N} = \mathrm{TransformerEncoder}\left(\bm{g}\right).
\end{align}
In this study, we set $N$ to $3$ for achieving a balance between computational efficiency and the model's ability to handle multiple references effectively.
Subsequently, we input the \verb|[CLS]| token $\bm{h}_{\texttt{[CLS]}}$ from $\bm{h}_\mathrm{N}$ into an MLP and then apply a sigmoid function to compute the final evaluation score $\hat{y}$.

We employ the Huber loss as our loss function due to its robustness against outliers. 
The Huber loss is defined using $\hat{y}$, the human judgments $y$, and the hyperparameter $\delta$ as follows:
\begin{align}
    \mathcal{L} = 
    \begin{cases} 
        \frac{1}{2}(\hat{y} - y)^{2} & \text{if } |\hat{y} - y| < \delta, \\
        \delta \cdot (|\hat{y} - y| -  \frac{1}{2}  \cdot \delta) & \text{otherwise.} 
    \end{cases}
\end{align}
In this study, we set $\delta$ to 0.5.

\section{Experiments and Results}
\label{sec:exp}

\subsection{Experimental Setup}

\paragraph{\textbf{Nebula dataset.}}
The development of supervised metrics for image captioning requires a large-scale, diverse dataset, but there are limited available datasets for effectively training these models. 
Moreover, the largest dataset in this field\cite{polos} has a notable issue: a significant imbalance, characterized by a limited variety of images relative to the number of captions. 
Specifically, there is a discrepancy where the number of images is approximately only one-tenth of the total number of samples.
This imbalance could potentially lead to suboptimal evaluation of various types of images.

To mitigate this imbalance, we constructed the Nebula dataset\footnote{The Nebula dataset can be downloaded from \href{https://deneb-project-page-nc03k.kinsta.page/}{this link}.} by extending the Polaris dataset to have approximately three times the number of images.
Following the standard procedure, human judgments were adopted on a five-point scale to assess the appropriateness of a candidate for a given image and references.
The annotation process was carried out through a crowdsourcing service.
Following previous studies \cite{umic, jaspice, polos}, we instructed the annotators to assess the quality of the candidates from the perspectives of fluency, relevance, and descriptiveness.
To ensure the reliability of our data, we excluded data from evaluators who exhibited suspicious behavior, such as extremely short response times or consistently providing identical values.
In addition, the human judgments, given on a five-point scale, were normalized to the range $[0,1]$.
For a comprehensive description of the dataset, please refer to the Appendix.

\subsection{Comparison to State-of-the-Art}

We evaluated the automatic evaluation metrics based on the accuracy they achieved when applied to the FOIL benchmark and PASCAL-50S, as well as their correlation coefficients when applied to Composite, Flickr8K-Expert, Flickr8K-CF, and Nebula. 
Given the importance of efficiency in practical applications, we also conducted experiments to measure the inference times of these metrics. This enables a comprehensive evaluation of both the performance effectiveness and practical usability of these metrics.

We adopted BLEU\cite{bleu}, ROUGE\cite{rouge}, METEOR\cite{meteor}, CIDEr\cite{cider}, and SPICE\cite{spice} as they are standard metrics for image captioning tasks. 
Additionally, we included MoverScore\cite{moverscore}, BERTScore\cite{bertscore}, BARTScore\cite{bartscore}, TIGEr\cite{tiger}, LEIC\cite{leic}, ViLBERTScore\cite{vilbertscore}, UMIC\cite{umic}, MID\cite{mid}, CLIP-S\cite{clipscore}, PAC-S\cite{pac-s}, Polos\cite{polos}, CLAIR\cite{clair} and FLEUR\footnote{Note that FLEUR and RefFLEUR was released after the submission of our manuscript. At the time of our submission, \textsc{Deneb} exhibited the best performance.} \cite{lee2024fleur} as baseline metrics since they are representative metrics for image captioning.
It is important to note that among these, CLAIR and FLEUR are LLM-based metrics, while the others are LLM-free.
\begin{table}[t]
    \centering
    \normalsize
    \caption{A quantitative comparison with baseline metrics. \textbf{Boldface} indicates the best values, and \underline{underlining} indicates the second best values. A ``-'' indicates either non-executable code or unavailable data.
    Note that, for a fair comparison, we have also included \textsc{Deneb} using OpenCLIP ViT-L/32 backbone as the encoder.
    }
    \setlength{\tabcolsep}{5pt}
    \scalebox{0.7}{
    \begin{tabular}{
    >{\raggedright\arraybackslash}p{45mm}
    >{\centering\arraybackslash}p{14mm}
    >{\centering\arraybackslash}p{14mm}
    >{\centering\arraybackslash}p{14mm}
    >{\centering\arraybackslash}p{14mm}
    >{\centering\arraybackslash}p{14mm}
    >{\centering\arraybackslash}p{14mm}
    >{\centering\arraybackslash}p{16mm}
    }
    \toprule
    {} & {\small \textbf{\begin{tabular}{c}FOIL\\1-ref\end{tabular}}} & {\small \textbf{\begin{tabular}{c}FOIL\\4-ref\end{tabular}}} & {\small \textbf{Composite}} & {\small \textbf{\begin{tabular}{c}Flickr8K\\Expert\end{tabular}}} & {\small \textbf{\begin{tabular}{c}Flickr8K\\CF\end{tabular}}} & {\small \textbf{Nebula}} & {\small \textbf{\begin{tabular}{c}Inference\\time\end{tabular}}} \\  
        \cmidrule(lr){2-3} \cmidrule(lr){4-7} \cmidrule(lr){8-8}
            {} & \multicolumn{2}{c}{\footnotesize Accuracy [\%]} & \multicolumn{4}{c}{\footnotesize Kendall's $\tau$} & {\footnotesize [ms]} \\
\cmidrule(r){1-1} \cmidrule(l){2-8} \rowcolor{TitleColor} \multicolumn{2}{l}{\footnotesize \textbf{\textit{LLM-free metrics}}} & {} & {} & {}& {} & {} & {} \\ 
{~~BLEU\cite{bleu}} & {66.5} & {82.6} & {30.6} & {30.8} & {16.4} & {40.4} & {0.13} \\ 
{~~ROUGE\cite{rouge}} & {71.7} & {79.3} & {32.4} & {32.3} & {19.9} & {42.6} & {0.51} \\
{~~CIDEr\cite{cider}} & {82.5} & {90.6} & {37.7} & {43.9} & {24.6} & {48.1} & {0.40} \\
{~~METEOR\cite{meteor}} & {78.8} & {82.6} & {38.9} & {41.8} & {22.2} & {46.8} & {0.14} \\
{~~SPICE\cite{spice}} & {75.5} & {86.1} & {40.3} & {44.9} & {24.4} & {44.0} & {17} \\
{~~UMIC\cite{umic}} & {--} & {--} & {56.1} & {46.8} & {30.1} & {--} & {--} \\
{~~CLIP-S\cite{clipscore}} & {87.2} & {87.2} & {53.8} & {51.2} & {34.4} & {46.9} & {1.7} \\
{~~PAC-S\cite{pac-s} (ViT-B/32)} & {89.9} & {89.9} & {55.7} & {54.3} & {36.0} & {47.2} & {2.5} \\
{~~PAC-S\cite{pac-s} (ViT-L/14)} & {93.1} & {93.1} & {54.0} & {55.7} & {37.0} & {47.9} & {12} \\
{~~SPARCS\cite{smurf}} & {--} & {--} & {43.1} & {48.1} & {10.4} & {46.5} & {1.5} \\ 
{~~BERTScore\cite{bertscore}} & {88.6} & {92.1} & {30.1} & {46.7} & {22.8} & {47.0} & {7.6} \\
{~~BARTScore\cite{bartscore}} & {85.3} & {91.1} & {43.5} & {37.8} & {24.3} & {43.8} & {130} \\
{~~ViLBERTScore\cite{vilbertscore}} & {--} & {--} & {52.4} & {50.1} & {--} & {--} & {--} \\
{~~MID\cite{mid}} & {90.5} & {90.5} & {55.7} & {54.9} & {37.3} & {51.3} & {52} \\
{~~RefCLIP-S\cite{clipscore}} & {91.0} & {92.6} & {55.4} & {53.0} & {36.4} & {46.9} & {2.8} \\
{~~RefPAC-S\cite{pac-s} (ViT-B/32)} & {93.7} & {94.9} & {57.3} & {55.9} & {37.6} & {50.6} & {4.5} \\
{~~RefPAC-S\cite{pac-s} (ViT-L/14)} & \underline{94.4} & {94.9} & {56.0} & \underline{56.4} & \underline{37.8} & {50.4} & {17} \\
{~~Polos\cite{polos}} & {93.2} & \underline{{95.1}} & \underline{{57.6}} & \underline{{56.4}} & \underline{{37.8}} & \underline{{53.9}} & {6.9} \\
\cmidrule(r){1-1} \cmidrule(l){2-8}
{} & \textbf{95.1} & \textbf{96.1} & \textbf{57.9} & \textbf{56.5} & \textbf{38.0} & \textbf{54.1} & 7.3 \\ 
\multirow{-2}{*}{~~\textbf{\textsc{Deneb} (ViT-B/32)}} & (\textcolor{blue}{+0.7}) & (\textcolor{blue}{+1.0}) & (\textcolor{blue}{+0.3}) & (\textcolor{blue}{+0.1}) & (\textcolor{blue}{+0.2}) & (\textcolor{blue}{+0.2}) & {} \\ 
{} & \textbf{95.4} & \textbf{96.5} & \textbf{58.2} & \textbf{56.8} & \textbf{38.3} & \textbf{54.3} & 22 \\ 
\multirow{-2}{*}{~~\textbf{\textsc{Deneb} (ViT-L/14)}} & (\textcolor{blue}{+1.0}) & (\textcolor{blue}{+1.4}) & (\textcolor{blue}{+0.6}) & (\textcolor{blue}{+0.4}) & (\textcolor{blue}{+0.5}) & (\textcolor{blue}{+0.4}) & {} \\ 
\cmidrule(r){1-1} \cmidrule(l){2-8}\rowcolor{TitleColor} \multicolumn{2}{l}{\footnotesize \textbf{\textit{LLM-based metrics}}} & {} & {} & {}& {} & {} & {} \\ 
{~~CLAIR\cite{clair}} & {81.4} & {83.4} & {55.0} & {44.6} & {34.4} & {52.7} & {1600} \\
{~~FLEUR\cite{lee2024fleur}} & \underline{96.8} & \underline{96.8} & \underline{63.5} & \textbf{53.0} & \underline{38.6} & {--} & {700} \\
{~~RefFLEUR\cite{lee2024fleur}} & \textbf{97.3} & \textbf{98.4} & \textbf{64.2} & \underline{51.9} & \textbf{38.8} & {--} & {760} \\
\bottomrule
    \end{tabular}
    }

    \label{tab:quantitative_result}

\end{table}

\paragraph{\textbf{Hallucination and Likert judgments.}}
We conducted comparative experiments against baseline models across the FOIL dataset to investigate the robustness against hallucinations. Additionally, we performed quantitative comparisons on various datasets, including Composite, Flickr8K-Expert, Flickr8K-CF, and Nebula.
Table \ref{tab:quantitative_result} shows the performance of the proposed and baseline metrics on FOIL.
Following previous studies \cite{polos, pac-s, clipscore}, our experiments were conducted in settings where either one or four reference captions were provided. 
In both single-reference (1-ref) and four-reference (4-ref) settings, our metric achieved state-of-the-art (SOTA) performance, with scores of 95.4\% and 96.4\%, respectively, marking improvements of 1.7 and 1.4 points over existing LLM-free metrics.
This suggests that our metric is more effective in addressing hallucinations compared to existing metrics.

Table \ref{tab:quantitative_result} also shows the quantitative comparison results with baseline metrics for the Composite, Flickr8K, Flickr8K-CF, and Nebula datasets. 
Consistent with previous studies \cite{clipscore, mid, spice, polos}, we employed $\tau_{b}$ (Kendall-B) for Flickr8K-CF and $\tau_{c}$ (Kendall-C) for the other datasets.
It should be noted that the Kendall-C evaluation values for the CLAIR metric were not provided in \cite{clair}; thus, for a fair comparison, we reproduced CLAIR and reevaluated it using Kendall-C. 
This illustrates that our proposed metric \textsc{Deneb} achieved SOTA results with scores of 58.2, 56.8, 38.3, and 54.3 on the Composite, Flickr8K-Expert, Flickr8K-CF, and Nebula datasets, respectively. 
In particular, \textsc{Deneb} outperformed the existing SOTA LLM-free metrics by margins of 0.6, 0.4, 0.5, and 0.4 on the Composite, Flickr8K-Expert, Flickr8K-CF, and Nebula datasets, respectively.

\paragraph{\textbf{Inference time.}}
In the context of practical applications, it is essential to consider not only the performance of metrics but also their inference times. 
Table \ref{tab:quantitative_result} shows the inference times per sample, measured on a system equipped with a GeForce RTX 3090 and an Intel Core i9-10900KF.
 The inference times for recent LLM-free metrics, RefCLIP-S and RefPAC-S, were 2.8 ms and 4.5 ms, respectively. 
Similarly, \textsc{Deneb} demonstrated an inference time of 7.3 ms, suggesting that the proposed metric is comparable in terms of inference speed. 
In contrast, LLM-based metrics such as CLAIR and FLEUR exhibited significantly longer inference times of 1600 ms and 700 ms respectively, which are at least 95 times slower than \textsc{Deneb}.
These results indicated the limitations of LLM-based metrics, rendering them impractical for real-time applications.
Considering that standard evaluation datasets sometimes contain more than 40,0000 samples \cite{flickr}, it is crucial that the evaluations are completed within a practical timeframe.

\begin{table}[t]
    \caption{A quantitative comparison on PASCAL-50S. \textsc{Deneb} achieved the best or second-best performance across all categories.}
    \centering
    \normalsize
    \scalebox{0.8}{
    \begin{tabular}{
    >{\raggedright\arraybackslash}p{50mm}
    >{\centering\arraybackslash}p{13mm}
    >{\centering\arraybackslash}p{13mm}
    >{\centering\arraybackslash}p{13mm}
    >{\centering\arraybackslash}p{13mm}
    >{\centering\arraybackslash}p{13mm}
    }
    \toprule
{} & {\textbf{HC}} & {\textbf{HI}} & {\textbf{HM}} & {\textbf{MM}} & {\textbf{Mean}} \\ 
\cmidrule(l){2-6}
{} & \multicolumn{5}{c}{\footnotesize Accuracy [\%]} \\ 
\cmidrule(r){1-1} \cmidrule(l){2-6}
\rowcolor{TitleColor} \multicolumn{2}{l}{\textbf{\textit{LLM-free metrics}}} & {} & {} & {} & {} \\ 
{~~~~BLEU\cite{bleu}} & {60.4} & {90.6} & {84.9} & {54.7} & {72.7} \\
{~~~~METEOR\cite{meteor}} & {63.8} & {97.7} & {93.7} & {65.4} & {80.2} \\
{~~~~ROUGE\cite{rouge}} & {63.7} & {95.3} & {92.3} & {61.2} & {78.1} \\
{~~~~SPICE\cite{spice}} & {63.6} & {96.3} & {86.7} & {68.3} & {78.7} \\
{~~~~CIDEr\cite{cider}} & {65.1} & {98.1} & {90.5} & {64.8} & {79.6} \\ 
{~~~~CLIP-S\cite{clipscore}} & {56.5} & {99.3} & {96.4} & {70.4} & {80.7} \\
{~~~~PAC-S\cite{pac-s} (ViT-B/32)} & {60.6} & {99.3} & {96.9} & {72.9} & {82.4} \\ 
{~~~~PAC-S\cite{pac-s} (ViT-L/14)} & {59.6} & \underline{99.7} & {96.9} & {75.2} & {82.9} \\ 
{~~~~UMIC\cite{umic}} & {66.1} & \textbf{99.8} & \textbf{98.1} & {76.2} & {85.1} \\ 
{~~~~BERTScore\cite{bertscore}} & {65.4} & {98.1} & {96.4} & {60.3} & {80.1} \\
{~~~~MoverScore\cite{moverscore}} & {65.1} & {97.1} & {93.2} & {65.6} & {80.3} \\
{~~~~TIGEr\cite{tiger}} & {56.0} & \textbf{99.8} & {92.8} & {74.2} & {80.7} \\
{~~~~MID\cite{mid}} & {67.0} & \underline{{99.7}} & \underline{{97.4}} & {76.8} & {85.2} \\ 
{~~~~RefCLIP-S\cite{clipscore}} & {64.5} & {99.6} & {95.4} & {72.8} & {83.1} \\
{~~~~RefPAC-S\cite{pac-s} (ViT-B/32)} & {67.7} & {99.6} & {96.0} & {75.6} & {84.7} \\ 
{~~~~RefPAC-S\cite{pac-s} (ViT-L/14)} & {65.0} & \textbf{99.8} & {97.3} & {76.1} & {84.6} \\ 
{~~~~Polos\cite{polos}} & \underline{{70.0}} & {99.6} & {\underline{97.4}} & \textbf{79.0} & \underline{86.5} \\
 \cmidrule(r){1-1} \cmidrule(l){2-6} 
{} & \textbf{74.4} & \textbf{99.8} & {97.3} & {76.5} & \textbf{87.0} \\ 
\multirow{-2}{*}{~~~~\textbf{\textsc{Deneb} (ViT-B/32)}} & (\textcolor{blue}{+4.4}) & {} & {} & {} & (\textcolor{blue}{+1.3}) \\ 
{} & \textbf{76.1} & \underline{{99.7}} & \underline{{97.4}} & \underline{77.9} & \textbf{87.8} \\ 
\multirow{-2}{*}{~~~~\textbf{\textsc{Deneb} (ViT-L/14)}} & (\textcolor{blue}{+6.1}) & {} & {} & {} & (\textcolor{blue}{+0.5}) \\ 
 \cmidrule(r){1-1} \cmidrule(l){2-6} 
\rowcolor{TitleColor} \multicolumn{2}{l}{\footnotesize \textbf{\textit{LLM-based metrics}}} & {} & {} & {}& {}\\ 
{~~~~CLAIR-E\cite{clair}} & {57.7} & \textbf{99.8} & {94.6} & \underline{75.6} & {81.9} \\ 
{~~~~FLEUR\cite{lee2024fleur}} & \underline{61.3} & \underline{99.7} & \underline{97.6} & {74.2} & \underline{83.2} \\
{~~~~RefFLEUR\cite{lee2024fleur}} & \textbf{68.0} & \textbf{99.8} & \textbf{98.0} & \textbf{76.1} & \textbf{85.5}\\
\bottomrule
    \end{tabular}
    }

    \label{tab:quantitative_result_pascal}
\end{table}

\paragraph{\textbf{Pairwise ranking on Pascal-50S.}}
PASCAL-50S consists of 1,000 images each with 50 reference captions, and presents the task of identifying which caption out of a given pair has received the majority vote from human judgments.
Specifically, this task focuses on pairwise preference judgments between two captions in the following categories:  pairs of HC (human correct) captions, HI pairs (both human-written, with one being incorrect), HM pairs (one from a human and the other machine-generated), and MM pairs (both generated by machines).
Table \ref{tab:quantitative_result_pascal} presents the performance of our proposed metric with baselines in PASCAL-50S.
As indicated in Table \ref{tab:quantitative_result_pascal}, \textsc{Deneb} achieved SOTA performance with scores of 76.1\% in HC, 77.9\% in MM, and Mean of 87.8\%, outperforming existing metrics by margins of 6.1, 0.6, and 1.3 points, respectively.
\begin{figure*}[t]
\scalebox{0.75}{
\begin{tikzpicture}
\newlength{\imageheight}
\settoheight{\imageheight}{\includegraphics[height=3cm, clip]{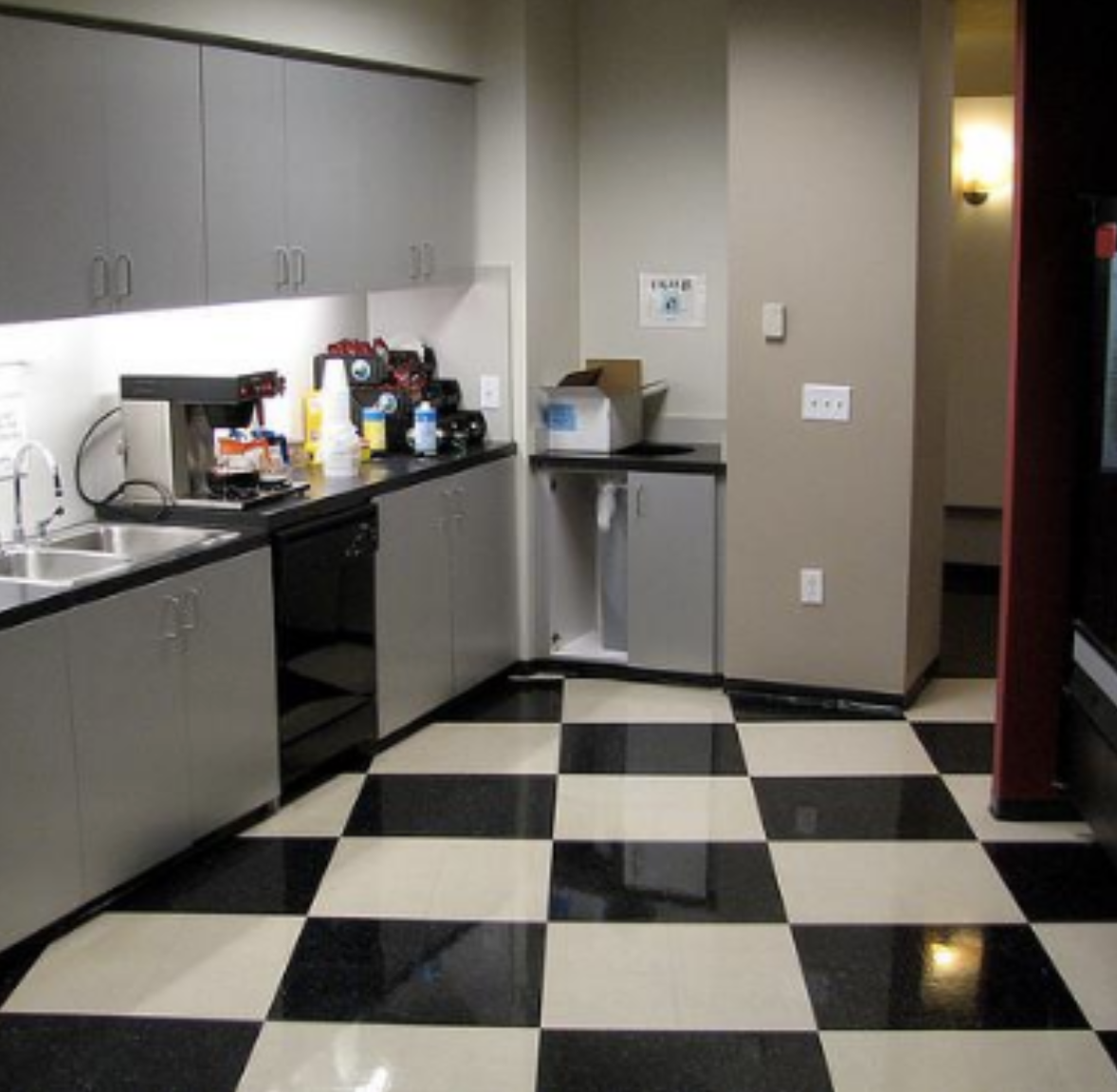}}
\node[inner sep=0pt] (image) at (0,0) {\includegraphics[height=3cm, width=3cm, clip]{fig/success1.png}};
\node[align=left, fill=white, text width=6cm, right=2mm of image, minimum height=\imageheight, inner sep=2pt] (desc) {
\large $\bm{x}_\mathrm{ref}^{(1)}$: ``A kitchen with vending machines and a black and white checkered floor'' \\
\textcolor{black}{\large $\bm{x}_\mathrm{cand}$: ``black and white checkered floor''} \\
\textbf{Human: 0.5}
};
\draw[black, thick] ([shift={(-1mm,-1mm)}]image.south west) rectangle ([shift={(1mm,1mm)}]desc.north east);
\node[align=center, draw=black, thick, inner sep=2pt, right=2mm of desc.east, text width=.3\textwidth, minimum height=\imageheight+8pt] (metrics) {
    \begin{tabular}{ccc}
    \textbf{\textcolor{black}{RefCLIP-S}} & \textbf{\textcolor{black}{Polos}} \\
    \textbf{\textcolor{black}{\large 0.719}} & \textbf{\textcolor{black}{\large 0.816}} \\
    \end{tabular}
};
\node[align=center, draw=mygreen, thick, inner sep=8pt, right=1mm of metrics, text width=.1\textwidth, minimum height=\imageheight+8pt] (metrics4) {
\textbf{\textcolor{mygreen}{\textsc{Deneb}\\\Large 0.634}}
};
\node[anchor=north west, xshift=-1.0cm] at (image.west) {\large (a)};
\end{tikzpicture}
}

\scalebox{0.75}{
\begin{tikzpicture}
\settoheight{\imageheight}{\includegraphics[height=3cm, clip]{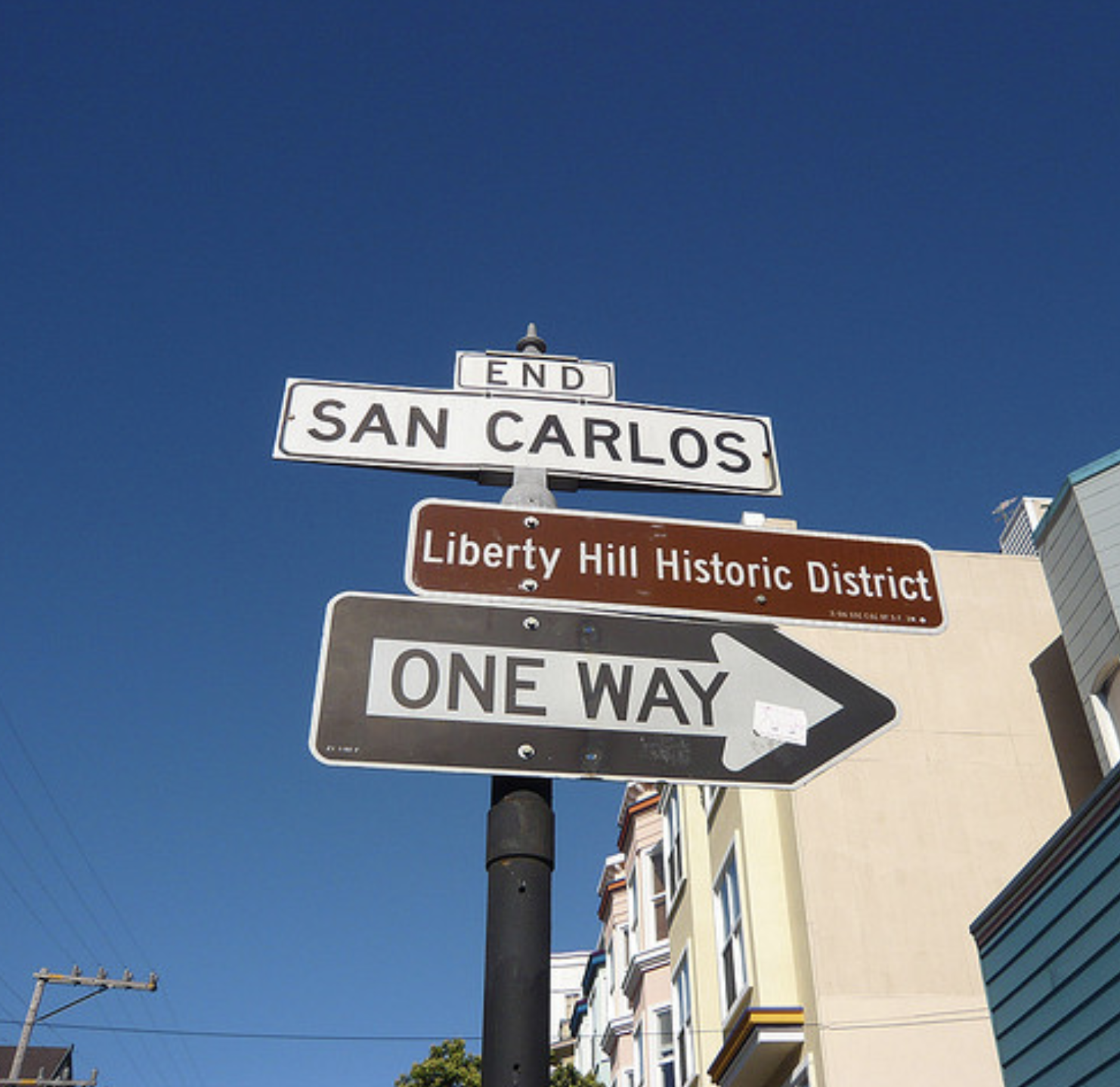}}
\node[inner sep=0pt] (image) at (0,0) {\includegraphics[height=3cm, width=3cm, clip]{fig/success2.png}};
\node[align=left, fill=white, text width=6cm, right=2mm of image, minimum height=\imageheight, inner sep=2pt] (desc) {
\large $\bm{x}_\mathrm{ref}^{(1)}$: ``Three traffic signs arranged on a sign post.'' \\
\textcolor{black}{\large $\bm{x}_\mathrm{cand}$: ``“a man sitting
on a stop sign on a street corner''} \\
\textbf{Human: 0.0}
};
\draw[black, thick] ([shift={(-1mm,-1mm)}]image.south west) rectangle ([shift={(1mm,1mm)}]desc.north east);
\node[align=center, draw=black, thick, inner sep=2pt, right=2mm of desc.east, text width=.3\textwidth, minimum height=\imageheight+8pt] (metrics) {
    \begin{tabular}{ccc}
    \textbf{\textcolor{black}{RefCLIP-S}} & \textbf{\textcolor{black}{Polos}}\\
    \textbf{\textcolor{black}{\large 0.442}} & \textbf{\textcolor{black}{\large 0.392}} \\
    \end{tabular}
};
\node[align=center, draw=mygreen, thick, inner sep=8pt, right=1mm of metrics, text width=.1\textwidth, minimum height=\imageheight+8pt] (metrics4) {
\textbf{\textcolor{mygreen}{\textsc{Deneb}\\\Large 0.023}}
};
\node[anchor=north west, xshift=-1.0cm] at (image.west) {\large (b)};
\end{tikzpicture}
}

\scalebox{0.75}{
\begin{tikzpicture}
\settoheight{\imageheight}{\includegraphics[height=3cm,clip]{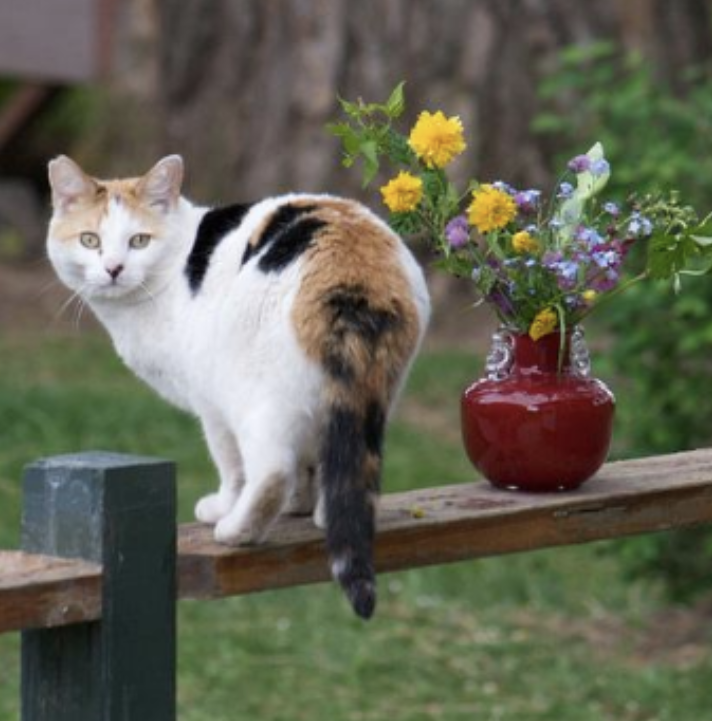}}
\node[inner sep=0pt] (image) at (0,0) {\includegraphics[height=3cm, width=3cm,clip]{fig/fail.png}};
\node[align=left, fill=white, text width=6cm, right=2mm of image, minimum height=\imageheight, inner sep=2pt] (desc) {
\large $\bm{x}_\mathrm{ref}^{(1)}$: ``A small cat standing on a wood railing next to an arrangement in a vase.'' \\
\textcolor{black}{\large $\bm{x}_\mathrm{cand}$: ``a cat on a fence''} \\
\textbf{Human: 0.25}
};
\draw[black, thick] ([shift={(-1mm,-1mm)}]image.south west) rectangle ([shift={(1mm,1mm)}]desc.north east);
\node[align=center, draw=black, thick, inner sep=2pt, right=2mm of desc.east, text width=.3\textwidth, minimum height=\imageheight+8pt] (metrics) {
    \begin{tabular}{ccc}
    \textbf{\textcolor{black}{RefCLIP-S}} & \textbf{\textcolor{black}{Polos}}  \\
    \textbf{\textcolor{black}{\large 0.591}} & \textbf{\textcolor{black}{\large 0.703}}  \\
    \end{tabular}
};
\node[align=center, draw=gray, thick, inner sep=8pt, right=1mm of metrics, text width=.1\textwidth, minimum height=\imageheight+8pt] (metrics4) {
\textbf{\textcolor{gray}{\textsc{Deneb}\\\Large 0.691}}
};
\node[anchor=north west, xshift=-1.0cm] at (image.west) {\large (c)};
\end{tikzpicture}
}
    \caption{Qualitative results on the Nebula dataset. Panels (a) and (b) illustrate successful cases, and panel (c) depicts a failure case.}
    \label{fig:qualitative_result}
\end{figure*}

\subsection{Qualitative Analysis}

Fig. \ref{fig:qualitative_result}-(a) shows a successful sample in the Nebula dataset.
In this sample, $\bm{x}_\mathrm{cand}$ is ``black and white checkered floor,'' and $\bm{x}_\mathrm{ref}^{(1)}$ is ``A kitchen with vending machines and a black and white checkered floor''.
The human judgment in this sample was rated 0.5 because $\bm{x}_\mathrm{cand}$ was partially correct.
While both Polos\cite{polos} and RefCLIP-S\cite{clipscore} incorrectly evaluated it as 0.816 and 0.719, respectively, \textsc{Deneb} evaluated it closer to $y$ with a score of 0.634.
Fig. \ref{fig:qualitative_result}-(b) shows another successful sample in the Nebula dataset.
In this sample, $\bm{x}_\mathrm{cand}$ is ``a man sitting on a stop sign on a street corner,'' and $\bm{x}_\mathrm{ref}^{(1)}$ is ``Three traffic signs arranged on a sign post.''
The human judgment for this sample was 0.0 due to the presence of hallucinations (``a man sitting on'') in $\bm{x}_\mathrm{cand}$.
Polos \cite{polos} and RefCLIP-S incorrectly evaluated this sample with scores of 0.392 and 0.442, respectively, whereas \textsc{Deneb} evaluated it more accurately with a value of 0.023.
These results demonstrate that our proposed metric \textsc{Deneb} is not only aligned more closely with human judgments but also shows robustness against hallucinations.

Fig. \ref{fig:qualitative_result}-(c) illustrates a failure sample in the Nebula dataset.
In this sample, $\bm{x}_\mathrm{cand}$ is described as ``a cat on a fence,'' while $\bm{x}_\mathrm{ref}^{(1)}$ is ``A small cat standing on a wood railing next to an arrangement in a vase.''
The human judgment for this sample was 0.25 due to the lack of sufficient detail.
However, \textsc{Deneb} inappropriately evaluated it with a score of 0.691.
Polos\cite{polos}, RefPAC-S\cite{pac-s}, and RefCLIP-S\cite{clipscore} also output significantly erroneous evaluation scores of 0.703, 0.847, and 0.591, respectively.
These consistent errors across metrics utilizing image features suggest that the failure of the proposed metric is likely due to an overestimation of the quality of $\bm{x}_\mathrm{cand}$, which only describes objects that are prominently displayed in the image. 

\subsection{Ablation Studies}

We conducted ablation studies to investigate the contribution of each module. 
Table \ref{tab:ablation} presents the quantitative results of the ablation studies. 

\paragraph{\textbf{Sim-Vec Transformer ablation.}} \:
We replaced the Sim-Vec Transformer with MLP to investigate the contribution of the Sim-Vec Transformer.
Table \ref{tab:ablation} shows that the accuracy for 1-ref and 4-ref on FOIL in Metric (i) were 76.2\% and 76.5\%, respectively, representing a decrease of 19.2 and 20.0 points compared to Metric (iv).
Table \ref{tab:ablation} also shows that the correlation coefficients for Composite, Flickr8K-Expert,Flickr8K-CF, and Nebula in Metric (i) were 37.7, 40.1, 25.1, and 48.1, respectively. 
In comparison to Metric (iv), these values have decreased significantly by 20.7, 16.1, 13.2, and 6.2 points. 
These results demonstrate that the SVE Transformer contributes to both performance and robustness against hallucinations.

\paragraph{\textbf{SVE ablation.}} \:
To analyze the contribution of the SVE, we simply used Eq. (\ref{eq:2}) as input instead of Eqs. (\ref{eq:3}) to (\ref{eq:6}).
Table \ref{tab:ablation} shows that the accuracy for 1-ref and 4-ref on FOIL in Metric (ii) were 84.3\% and 89.3\%, respectively, representing a decrease of 11.1 and 7.2 points compared to Metric (iv).
Table \ref{tab:ablation} also shows that the correlation coefficients for Composite, Flickr8K-Expert, Flickr8K-CF, and Nebula in Metric (ii) were 35.8, 49.6, 24.7, and 45.2, respectively. 
These values were lower by 22.6, 6.6, 13.6, and 9.1 points than Metric (iv).
These results indicate that the SVE is effective in extracting significant features for automatic evaluation and enhancing robustness against hallucinations.

\paragraph{\textbf{Multifaceted references ablation.}} \:
We did not originally employ Aggregate functions in \textsc{Deneb} to handle multifaceted references, as detailed in Section \ref{sec:method}. 
To explore their effect on performance, we conducted an ablation study by adding Aggregate functions. 
Table \ref{tab:ablation} shows that the accuracy for 1-ref and 4-ref on FOIL in Metric (iii) were 94.4\% and 96.1\%, respectively, representing a decrease of 1.0 and 0.4 points compared to Metric (iv).
Table \ref{tab:ablation} also demonstrates that the correlation coefficients for Composite, Flickr8K-Expert, Flickr8K-CF, and Nebula in Metric (iii) were 57.4, 55.7, 37.2, and 53.2, respectively. 
In comparison to Metric (iv), these values were lower by 1.0, 0.5, 1.1, and 1.1 points.
These results indicate that the inclusion of a mechanism for handling multifaceted references contributed to both performance and robustness against hallucinations.
\begin{table}[t]
    \caption{Quantitative results of the ablation studies. The incorporation of the SVE Transformer, as indicated for Metrics (i) and (ii), emerges as the most significant factor in enhancing performance.}
    \centering
    \normalsize
    \setlength{\tabcolsep}{2pt}
    \renewcommand{\arraystretch}{1.2}
    \scalebox{1.0}{
    \begin{tabular}{cccccccccc}
    \toprule
{} & {\scriptsize \begin{tabular}[c]{@{}c@{}}Sim-Vec \\ Transformer\end{tabular}} & {\scriptsize SVE} & {\scriptsize \begin{tabular}[c]{@{}c@{}}Multifaceted \\ references\end{tabular}} & {\scriptsize \begin{tabular}[c]{@{}c@{}}FOIL \\ 1-ref\end{tabular}} & {\scriptsize \begin{tabular}[c]{@{}c@{}}FOIL \\ 4-ref\end{tabular}} & {\scriptsize \begin{tabular}[c]{@{}c@{}}Flickr8K \\ CF\end{tabular}} & {\scriptsize \begin{tabular}[c]{@{}c@{}}Flickr8K \\ Expert\end{tabular}} & {\scriptsize Composite} & {\scriptsize Nebula}  \\ 
\cmidrule(lr){5-6} \cmidrule(lr){7-10}
{} & {} & {} & {} & \multicolumn{2}{c}{\scriptsize Accuracy [\%]} & \multicolumn{4}{c}{\scriptsize Kendall's $\tau$} \\
\cmidrule(r){1-4} \cmidrule(l){5-10}   
{(i)} & {} & {} & {} & {76.2} & {76.5} & {25.1} & {40.1} & {37.7} & {48.1}\\
{(ii)} & {$\checkmark$} & {} & {$\checkmark$} & {84.3} & {89.3} & {24.7} & {49.6} & {35.8} & {45.2} \\
{(iii)} & {$\checkmark$} & {$\checkmark$} & {} & {94.4} & {96.1} & {37.2} & {55.7} & {57.4} & {53.2} \\ 
{(iv)} & {$\checkmark$} & {$\checkmark$} & {$\checkmark$} & \textbf{95.4} & \textbf{96.5}  & \textbf{38.3} & \textbf{56.2} & \textbf{58.4} & \textbf{54.3} \\ \bottomrule
    \end{tabular}
    }
    \label{tab:ablation}
\end{table}
\section{Limitations and Discussion}
While our metric has clearly been shown to provide impressive robustness against hallucinations, it is not without its limitations. The primary limitation is the occurrence of errors stemming from differences in the areas of focus. Moreover, the metric tends to erroneously overevaluate candidates that only describe prominent objects in an image while neglecting finer details. These limitations indicate that the proposed metric may not effectively capture the relationship between $x_{\text{cand}}$ and the local regions of $x_{\text{img}}$.
For further error analysis, see Appendix.

\section{Conclusions}
In this study, we have proposed the novel automatic evaluation metric \textsc{Deneb}, which is specifically robust to hallucinations.
Our key contributions are as follows: 
(1) We have introduced the Sim-Vec Transformer, which handles the similarity between an image, a candidate caption, and reference captions. 
(2) We proposed the Sim-Vec Extraction (SVE), which utilizes a Hadamard product and element-wise difference to extract features beneficial for automatic evaluation. 
(3) We constructed the diverse and balanced Nebula dataset comprising 32,978 images, paired with human judgments provided by 805 annotators. 
(4)We achieved state-of-the-art performance on FOIL, Composite, Flickr8K-Expert, Flickr8K-CF, PASCAL-50S, and the Nebula dataset.

\subsubsection*{Acknowledgements}
This work was partially supported by JSPS KAKENHI Grant Number 23H03478, JST CREST, and NEDO.

\bibliographystyle{splncs04}
\bibliography{main}

\appendix
\renewcommand{\thefigure}{\Alph{figure}}
\renewcommand{\thetable}{\Alph{table}}
\setcounter{figure}{0}
\setcounter{table}{0}
\title{\textsc{Deneb}: A Hallucination-Robust Automatic Evaluation Metric for Image Captioning\\(Supplementary Material)}
\author{Kazuki Matsuda \and Yuiga Wada \and Komei Sugiura}
\institute{Keio University, Japan\\\email{\{k2matsuda0, yuiga, komei.sugiura\}@keio.jp}}
\authorrunning{K.~Matsuda et al.}
\titlerunning{\textsc{Deneb}}

\maketitle
\section{Additional Related Work}
\label{appendix:additional_related_work}
The FOIL dataset, specifically designed to assess metric robustness against hallucination, is derived from the COCO dataset and comprises approximately 200,000 image-caption pairs with a mix of correct and hallucinated captions.
Flickr8K-Expert \cite{flickr}, Flickr8K-CF, Composite\cite{composite}, and Polaris\cite{polos} each consist of human Likert-scale judgments at the level of each image-caption pair. 
Specifically, Flickr 8K-Expert comprises 17,000 human judgments across 5,664 images, with captions rated on a four-point scale. 
Flickr8K-CF offers 145,000 human judgments collected from CrowdFlower, covering over 48,000 image-caption pairs. 
The Composite dataset, sourcing image-caption pairs from COCO\cite{coco}, Flickr8K \cite{flickr}, and Flickr30K \cite{flickr30k}, contains approximately 12,000 human judgments. 
The Polaris dataset is tailored for the training and evaluation of metrics and consists of a diverse collection of captions, sourced from ten modern image captioning models and accompanied by approximately 130,000 human judgments. 

\section{Implementation Details}
\label{appendix:implementation}
We divided the Nebula dataset into training, validation, and test sets, containing 26,382, 3,298, and 3,298 samples, respectively.
We used the training set to train our model, the validation set for hyperparameter tuning, and the test set for evaluating the model's performance.

Table \ref{tab:settings} shows the experimental settings of the proposed metric.
We employed early stopping for model training with a focus on Kendall's $\tau$.
This process entailed monitoring Kendall's $\tau$ on the validation set at each epoch.  
The training process was halted if Kendall’s $\tau$ on the validation set did not show any improvement for a single epoch.
Subsequently, we evaluated the model's performance using the test set.
\begin{table}[H]
    \normalsize
    \renewcommand*{\arraystretch}{1.25}
    \newcommand*{\bhline}[1]{\noalign{\hrule height #1}}
    \caption{The experimental settings for the proposed metric.}
    \centering
    \scalebox{0.95}{
    \begin{tabular}{lc}
    \toprule
        Optimizer & Adam ($\beta_{1}=0.9$, $\beta_{2}=0.999$) \\ \hline
        Learning rate & $5.0 \times 10^{-6}$ \\ \hline
        Batch size & 16 \\ \hline
        Loss function & Huber loss ($\delta=0.5$) \\
    \bottomrule
    \end{tabular}
    }
    \label{tab:settings}
\end{table}
Our model had approximately 133 million trainable parameters.
We trained our model on a Tesla A100 GPU and measured the inference time on the GeForce RTX 3090 with 24 GB of memory and an Intel Core i9 12900K with 64 GB of memory.
The training phase was completed in approximately 2 hours, and the inference time per sample on GeForce RTX 3090 was approximately 22 ms.

\section{Nebula Dataset}
To construct the Nebula dataset, we employed ten standard image captioning models.
These models include: $\mathrm{SAT}$ \cite{sat}, $\mathrm{\mathcal{M}^2}$-$\mathrm{Transformer}$ \cite{m2trm}, $\mathrm{VinVL}$ \cite{vinvl}, $\mathrm{GRIT}$ \cite{grit}, $\mathrm{BLIP_\mathrm{base}}$, $\mathrm{BLIP_\mathrm{large}}$ \cite{blip}, $\mathrm{GIT}$ \cite{git}, $\mathrm{OFA}$ \cite{ofa}, $\mathrm{BLIP}$-$\mathrm{2_\mathrm{flan}}$, and $\mathrm{BLIP}$-$\mathrm{2_\mathrm{opt}}$ \cite{blip2}. 
Here, $\mathrm{BLIP_\mathrm{base}}$ and $\mathrm{BLIP_\mathrm{large}}$ represent variants of BLIP that utilizes ViT-B and ViT-L\cite{ViT}, respectively.
$\mathrm{BLIP}$-$\mathrm{2_\mathrm{flan}}$ and $\mathrm{BLIP}$-$\mathrm{2_\mathrm{opt}}$ are variants of BLIP-2 that employs Flan-T5\cite{flant5} and OPT\cite{opt} as their large language models, respectively.

The images in the Nebula dataset were sourced from the validation sets of the MS-COCO \cite{coco} and nocaps \cite{nocaps} datasets. 
MS-COCO was selected as it is a standard dataset for image captioning, whereas nocaps was selected for its greater diversity of classes compared to MS-COCO. 
The validation sets of MS-COCO and nocaps were chosen to avoid potential data leakage that could occur when using their training sets, particularly in terms of evaluating an image captioning model trained on MS-COCO. 
Furthermore, their test sets were not used because they lacked the reference captions necessary for multifaceted metrics.

We instructed the annotators to assess the quality of the captions from the perspectives of fluency, relevance, and descriptiveness.
For fluency, they assessed the grammatical correctness of captions, deducting points for each grammatical error. 
For relevance, they evaluated whether the caption was closely related to the image and deducted points for irrelevant words. 
For descriptiveness, they assessed how comprehensively and accurately the caption describes the content of the image.

The Nebula dataset comprises 32,978 images and 32,978 human judgments collected from 805 annotators, and contains approximately three times more images than the Polaris dataset.
The total number of references is 183,472, with a vocabulary size of 32,870, a total word count of 1,945,956, and an average sentence length of 10.61 words. 
The total number of candidates is 32,978, with a vocabulary size of 3,695, a total word count of 288,922, and an average sentence length of 8.76 words. 
All sentences are in English.

\section{Additional Qualitative Analysis}
\label{appendix:additional}

Figs. \ref{fig:example1} and \ref{fig:example2} present additional qualitative results from the Nebula and FOIL datasets, respectively. 
In our analysis, we compared the performance of the \textsc{Deneb} model with three representative metrics: CIDEr \cite{cider} (\textit{classic}), CLIP-S \cite{clipscore} (\textit{reference-free}), and Polos \cite{polos} (\textit{pseudo-multifaceted}). 
Specifically, these methods have a tendency to overestimate the quality of instances where $\bm{x}_\mathrm{cand}$ were inappropriate but contained words related to the image. 
This discrepancy primarily stems from their limited capability to effectively compare candidates with multifaceted references and their significant reliance on the alignment of image and language features. 
In contrast, \textsc{Deneb} consistently assigned low evaluation scores to such captions and aligned closely with human judgments.
These results show its effectiveness and robustness against hallucinations.
\input{additional_examples_suppl}

\section{Error Analysis}
\label{appendix:error}

\begin{table}[t]
    \normalsize
    \renewcommand*{\arraystretch}{1.25}
    \newcommand*{\bhline}[1]{\noalign{\hrule height #1}}
    \caption{Categorization of failed samples.}
    \centering
    \scalebox{0.9}{
    \begin{tabular}{ m{0.55\linewidth} c}
    \bhline{0.6pt}
        Error Type & \#Error \\ \cmidrule(r){1-1} \cmidrule(l){2-2}
        Focus Area Discrepancy & 40 \\
        Caption Accuracy Deficiency  & 28 \\
        Caption Detail Insufficiency  & 16 \\
        Grammatical Error & 8 \\
        Annotation Error& 4 \\
        Others & 4 \\
    \bhline{0.6pt}
    \end{tabular}
    }
    \label{tab:error}
\end{table}

To investigate the limitations of the proposed method, we analyzed the worst 100 samples with the largest absolute differences between $\hat{y}$ and $y$.
We defined samples that satisfy $|y-\hat{y}| > 0.25$ as failure cases. Within the test set of Nebula dataset, a total of 503 samples were identified as failure cases.

Table \ref{tab:error} categorizes the failure cases. The causes of failure can be grouped into six main categories:
\begin{itemize}
    \setlength{\parskip}{0.5mm}
    \setlength{\itemsep}{0.2mm}
    \item[$\bullet$] Focus Area Discrepancy: This category pertains to samples where our metric incorrectly scores captions that focus on different areas than the references.
    \item[$\bullet$] Caption Accuracy Deficiency: This category refers to samples where our metric inappropriately scores captions with incorrect expressions.
    \item[$\bullet$] Caption Detail Insufficiency: This category pertains to samples where our metric outputs inappropriate scores when the candidate lacks details.
    \item[$\bullet$] Grammatical Error: This category refers to samples where our metric outputs inappropriate scores for captions containing grammatical errors.
    \item[$\bullet$] Annotation Error: This category includes samples where the human judgment was inappropriate.
    \item[$\bullet$] Others: This category covers other types of errors that do not fall into the aforementioned categories
\end{itemize}

Table \ref{tab:error} shows that the main bottleneck was errors due to differences in areas of focus.
In samples corresponding to these errors, $\bm{x}_\mathrm{cand}$ describes elements absent in $\bm{x}_\mathrm{ref}^{(1)}$, suggesting that the proposed metric may not effectively capture the relationship between $\bm{x}_\mathrm{cand}$ and the local region of $\bm{x}_\mathrm{img}$.
Consequently, the introduction of a mechanism to extract the relationship between features in local image regions and language features, as suggested in \cite{region-clip}, is anticipated to offer a possible solution.
In future work, we plan to extend \textsc{Deneb} by introducing a mechanism to extract the relationship between features in local image regions and language features, as suggested in \cite{region-clip}.

In future work, we plan to extend \textsc{Deneb} by introducing a mechanism to extract the relationship between features in local image regions and language features, as suggested in \cite{region-clip}.
\end{document}